\documentclass{article}

\PassOptionsToPackage{numbers, compress}{natbib}

\usepackage[final]{DSA_2025}

\usepackage{xcolor} 
\usepackage{enumitem} 
\usepackage[many]{tcolorbox}
\usepackage{boxedminipage}
\usepackage{graphicx}
\usepackage[utf8]{inputenc} 
\usepackage[T1]{fontenc}    
\usepackage{hyperref}       
\usepackage{url}            
\usepackage{booktabs}       
\usepackage{amsfonts}       
\usepackage{nicefrac}       
\usepackage{microtype}      
\usepackage{xcolor}         

\title{AI-Driven Climate Policy Scenario Generation for Sub-Saharan Africa}

\author{%
  Rafiu Adekoya Badekale\thanks{Code: The code for this paper can be found at  https://github.com/AdeTheBade/CPSG.git} \\
  Hamoye Foundation\\
  Delaware, United States \\
  \texttt{ade.badekale@hamoye.org} \\
  \And
  Adewale Akinfaderin \\
  The George Washington University \\
  Washington, D.C., United States \\
  \texttt{wale@hamoye.org} \\
}

\begin{document}

\maketitle

\begin{abstract}
Climate policy scenario generation and evaluation have traditionally relied on integrated assessment models (IAMs) and expert-driven qualitative analysis. These methods enable stakeholders, such as policymakers and researchers, to anticipate impacts, plan governance strategies, and develop mitigation measures. However, traditional methods are often time-intensive, reliant on simple extrapolations of past trends, and limited in capturing the complex and interconnected nature of energy and climate issues. With the advent of artificial intelligence (AI), particularly generative AI models trained on vast datasets, these limitations can be addressed, ensuring robustness even under limited data conditions. In this work, we explore the novel method that employs generative AI, specifically large language models (LLMs), to simulate climate policy scenarios for Sub-Saharan Africa. These scenarios focus on energy transition themes derived from the historical United Nations Climate Change Conference (COP) documents. By leveraging generative models, the project aims to create plausible and diverse policy scenarios that align with regional climate goals and energy challenges. Given limited access to human evaluators, automated techniques were employed for scenario evaluation. We generated policy scenarios using the llama3.2-3B model. Of the 34 generated responses, 30 (88\%) passed expert validation, accurately reflecting the intended impacts provided in the corresponding prompts. We compared these validated responses against assessments from a human climate expert and two additional LLMs (gemma2-2B and mistral-7B). Evaluation results show the model’s strong performance in generating factually accurate (faithfulness: 0.760 for human, 0.966 for gemma2-2B, and 0.848 for mistral-7B), highly relevant (answer relevancy: 0.993 for human, 0.888 for gemma2-2B, and 0.893 for mistral-7B), and well-contextualized (context utilization: 0.713 for human, 0.939 for gemma2-2B, and 0.980 for mistral-7B) scenarios. Spearman correlation analysis between human and model evaluations confirms a high degree of alignment, highlighting the reliability of automated evaluation. Our structured, embedding-based evaluation framework shows that generative AI effectively generate scenarios that are coherent, relevant, plausible, and diverse. This approach offers a transformative tool for climate policy planning in data-constrained regions.
\end{abstract}

\section{Introduction}
Sub-Saharan Africa (SSA) faces unique climate challenges, marked by high exposure to risks and limited access to high-quality data for climate planning \cite{pörtner2022climate}. Traditional climate policy scenario generation methods, such as IAMs (e.g., DICE, RICE) are often expert-driven and limited by scalability issues, particularly in modeling complex regional dynamics \cite{zhang2022aiglobalclimatecooperation}. These models often fall short in the SSA context due to their reliance on detailed, country-level emissions and economic data \cite{gambhir2019review}. For example, simulating the impact of a carbon tax in SSA requires granular emissions data, which is unavailable for many SSA countries due to capacity constraints and irregular reporting. To address these limitations, our main contributions are summarized as follows,

\begin{itemize}

\item We introduce a novel application of generative AI in policy scenario generation for climate and energy in Sub-Saharan Africa. 

\item We develop a new framework for evaluating generated scenarios using retrieval and generation metrics. 

\end{itemize}

The emergence of generative AI, particularly large language models (LLMs)\footnote{A large language model (LLM) is an artificial intelligence system trained on huge amounts of text to understand and generate human-like language. It can answer questions, summarize documents, or write content based on a given prompt.}, offers a transformative solution by synthesizing diverse policy scenarios from historical data. This study leverages retrieval-augmented generation (RAG)\footnote{Retrieval-Augmented Generation (RAG) is a method where the AI combines external information (retrieved from documents) with its own language generation to produce more accurate and informed responses. It helps the AI "look things up" before answering.} to simulate policy scenarios, rooted in the United Nations (UN) Climate Change Conference (COP) documents as a knowledge base. We validated RAG’s effectiveness in generating plausible region-specific scenarios and evaluated the quality of the generated responses using a human climate expert and two LLMs. Our approach aims to enhance the scalability, transparency, and relevance of scenario generation, focusing on energy transition challenges critical to the SSA region.

\section{Methodology}

\subsection{Data collection and preprocessing}

We collected 94 COP documents from the United Nations Framework Convention on Climate Change (UNFCCC) website, covering all conference meetings (convention, kyoto protocol, and paris agreement). We organized documents by meeting and processed them using LangChain’s PyPDFLoader. Preprocessing involved chunking with RecursiveCharacterTextSplitter (chunk size: 1000, overlap: 100) and embedding stored in a chroma vector database using ollama embeddings (model="nomic-embed-text:latest"). Quality control included manual verification and automated filtering to remove duplicates.

\begin{figure}[ht]
  \centering
  \includegraphics[width=0.9\textwidth]{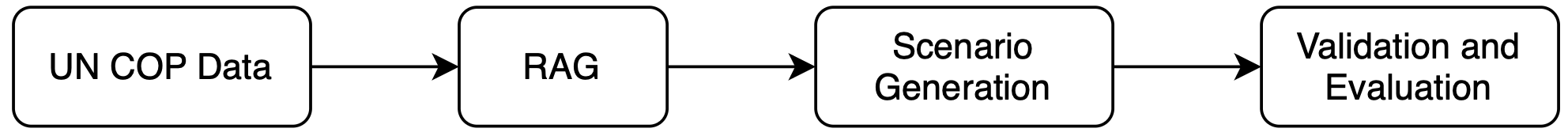}
  
  \vspace{1em} 
  \caption{Flow diagram of study}
  \vspace{1em} 
\end{figure}

\subsection{Scenario generation}

We used llama3.2-3B to generate scenarios with 34 zero-shot prompts, via a RAG pipeline grounded in the UN COP documents. The prompts were crafted from themes such as renewable energy adoption and technology transfer, identified via the same RAG setup. We set temperature to zero for consistency. To ground the generated scenarios in actionable steps (e.g., KPIs, milestones), we employed llama3-8B to prompt additional questions. 

\textbf{Example prompt:}

\textit{"You are a climate policy expert tasked with generating innovative and actionable future climate policy scenarios. Your responses should be relevant, detailed, and well-structured.}

\textit{Generate potential future climate policy scenarios for Sub-Saharan Africa that focus on accelerating renewable energy adoption. Consider policy incentives, grid integration challenges, and mechanisms for ensuring equitable access to clean energy.”}

\textbf{Example follow-up prompt:}

\textit{“Outline a clear implementation roadmap for each of the generated scenarios with well defined milestones. Define short-term (0-2 years), medium-term (3-7 years), and long-term (8+ years) milestones. Highlight critical decision points, dependencies, and risks that could impact progress. Provide specific action steps that should be taken at each milestone to ensure scenario success.”}

Having generated the model responses, we removed any extraneous text that did not align with the structure of the scenarios. This includes section headers and bullet points, which were sometimes added at the beginning or end of the generated responses. The main body of the generated scenarios remained intact and was not subject to manual or automated corrections of any type. 

\subsection{Scenario validation and evaluation}

Two authors independently validated the generated scenarios. Out of 34 generated responses, 30 (88\%) were deemed valid (highly depicting designated impacts provided in the corresponding prompt). We evaluated the quality of the validated responses using a three-metric Retrieval-Augmented Generation Assessment (RAGAs)\cite{es2023ragasautomatedevaluationretrieval} framework. These metrics were defined without relying on ground-truth human annotations, which was suitable for our case, as future scenarios are speculative rather than definitive. \textbf{Faithfulness} which refers to the idea that the answer should be grounded in the given context to avoid hallucinations. \textbf{Answer relevancy} which refers to the idea that the generated answer should address the actual prompt that was provided. Finally, \textbf{Context utilization} which refers to the idea that the retrieved context should be focused, containing as little irrelevant information as possible. Evaluators included a human climate expert and two LLMs (gemma2-2B, mistral-7B). Performance was analyzed using the mean scores for each evaluator across the 30 validated responses for the three metrics. To validate that the mean scores are representative of each evaluator’s performance, we computed the standard deviation (SD) as a measure of score variability. A low SD relative to the mean indicates that the scores are closely clustered and that the mean is a reliable summary statistic. Agreement between evaluators (inter-model and model-human) was assessed using spearman’s correlation coefficients.\footnote{Spearman correlation is a statistic used to measure how well the relationship between two sets of rankings matches. A score close to 1 means the rankings are very similar, while a score near 0 means they’re not.} The correlation was chosen for its ability in capturing monotonic relationships in ordinal, Likert-style scores without assuming linearity.

\section{Result}

\subsection{Evaluation metrics}

As shown in Table \ref{evaluation-table}, mean scores range from 0 to 1, with higher values indicating better performance. Answer relevancy exhibited the most consistent scoring among evaluators, as indicated by the SDs. The SDs for human, gemma2-2B, and mistral-7B were all below 0.060. This is less than 6\% of their respective means, indicating that the mean scores are highly representative for this metric. For faithfulness, gemma2-2B and human showed SDs below 11\% of their means, suggesting a reasonable level of scoring consistency. However, mistral-7B exhibited greater variability (approximately 27\% of the mean), indicating that its mean faithfulness score is less representative due to inconsistency across responses. Regarding context utilization, mistral-7B and human had SDs below 11\% of their respective means, supporting the use of the mean as a reliable summary. In contrast, gemma2-2B had a higher variability (approximately 19\% of the mean), indicating moderate inconsistency in scoring. Notably, among all evaluators, the human expert delivered the most consistent scoring across the three metrics. Overall, the mean scores are generally representative, particularly for answer relevancy. However, caution is advised when interpreting metrics like mistral-7B’s faithfulness and gemma2-2B’s context utilization, which exhibit higher variability.

Having established the reliability of the means, we noted that the evaluators produced consistently high answer relevancy scores. However, variation was observed in faithfulness and context utilization. For faithfulness, gemma2-2B achieved the highest mean score (0.966), followed by mistral-7B (0.848) and human (0.760). For context utilization, mistral-7B had the highest mean (0.980), followed by gemma2-2B (0.939) and human (0.713). Interestingly, gemma2-2B achieved the highest faithfulness score (0.966) but the lowest answer relevancy (0.888) among the evaluators. This counter-intuitive result reflects the evaluation behavior of gemma2-2B. The model appears to favor responses that adhere strictly to the retrieved context over directly addressing the prompt’s intent—i.e., evaluations that reward caution over directness. This trade-off highlights a potential tension between grounding and relevance in zero-shot settings for automated evaluation of scenario generation.

\begin{table}[ht]
  \caption{Mean scores ($\pm$ SD) for the evaluators across the validated generated responses}
  \label{evaluation-table}
  \centering
  \begin{tabular}{lccc}
    \toprule
    Evaluator         & Faithfulness       & Answer Relevancy   & Context Utilization \\
    \midrule
    Human      & 0.760 $\pm$ 0.081  & 0.993 $\pm$ 0.037  & 0.713 $\pm$ 0.101   \\
    \texttt{gemma2-2B} & 0.966 $\pm$ 0.102  & 0.888 $\pm$ 0.057  & 0.939 $\pm$ 0.184   \\
    \texttt{mistral-7B} & 0.848 $\pm$ 0.271  & 0.893 $\pm$ 0.050  & 0.980 $\pm$ 0.108   \\
    \bottomrule
  \end{tabular}
\end{table}

\begin{table}[ht]
  \caption{Spearman correlation coefficients between evaluator pairs}
  \label{correlation-table}
  \centering
  \begin{tabular}{lccc}
    \toprule
    Evaluator Pair       & Faithfulness & Answer Relevancy & Context Utilization \\
    \midrule
    \texttt{Mistral} vs \texttt{gemma2}   & 0.911        & 0.701            & 0.901               \\
    Human vs \texttt{gemma2}      & 0.814        & 0.311            & 0.859               \\
    Human vs \texttt{mistral}     & 0.919        & 0.311            & 0.401               \\
    \bottomrule
  \end{tabular}
\end{table}

\subsection{Correlation analysis}

Spearman correlation results (Table \ref{correlation-table}) show strong alignment between LLMs and the human evaluator for faithfulness. However, alignment is weak for answer relevancy and mixed for context utilization. These discrepancies highlight the challenges of LLMs in capturing human judgment.

\section{Discussion}

Our generative AI approach shows potential for scenario generation in resource-constrained regions \cite{jungwirth2023ai}. The RAG pipeline effectively leverages UN COP documents as a knowledge base, with llama3.2-3B demonstrating robust in-context learning capabilities. As shown in Table \ref{evaluation-table}, all evaluators assigned scores above 70\% to the responses generated by llama3.2-3B across all metrics. This indicates broad agreement on the overall quality of the generated responses. However, the correlation results in Table \ref{correlation-table} show that while LLMs align closely with humans in assessing factual consistency, gaps remain in interpreting answer relevancy. This suggests that LLMs struggle with subjective and contextual assessments, necessitating human oversight. This highlights the importance of our hybrid evaluation methods combining human judgment with automated tools. Limitations include reliance on zero-shot prompting, which may limit response depth, and potential biases from the base weights of LLMs. LLMs are commonly trained on general-purpose corpora dominated by western sources, which risks introducing a western skew \cite{wei2022chain}. This can result in cultural and contextual biases that misalign generated scenarios with the unique climate and energy priorities of SSA. Future work should explore advanced prompt engineering, ensemble models, and expanded datasets (e.g., national climate reports) to enhance accuracy and address biases. Future work could also fine-tune LLMs on region-specific corpora to better reflect local perspectives and values. Addressing representational biases and improving model alignment with human judgment are critical for ethical deployment in policy-making contexts \cite{ouyang2022training}.

For policymakers, this generative AI approach presents many practical applications. For instance, it could support a Ministry of Energy scenario workshop to explore energy transition pathways tailored to SSA's unique challenges. It could also facilitate a regional ECOWAS planning exercise to coordinate climate strategies across West African states, informed by data-driven narratives. By improving the accessibility and relevance of policy scenarios, this approach can foster more inclusive and informed climate governance in low-resource settings.

\section{Conclusion}

Our method provides a scalable, AI-driven framework for generating diverse and coherent climate policy scenarios tailored to SSA. The RAG-based approach, grounded in UN COP documents and the integration of human and automated evaluations confirms the robustness of our approach. While limitations in model alignment and biases require further research, this work provides a foundation for AI-driven policy planning, with broader applications in global climate research.

\begin{ack}
This work is fully funded by Hamoye Foundation, a 501(c)(3) organization.
\end{ack}


\newpage

\section*{Appendix}
 

\appendix

This section provides additional details to support the main findings of the paper, enhancing transparency and reproducibility. The appendices include the full prompt list (Appendix A), sample generated responses with evaluator notes (Appendix B) and the evaluation criteria used by the human climate expert (Appendix C). The code for the scenario generation and evaluation, along with the UN COP dataset, is available at [GitHub link: https://github.com/AdeTheBade/CPSG.git] for reproducibility.

\section{Validated prompts}
\label{app:prompts}

\begin{tcolorbox}[
    colback=lightgray, 
    colframe=yellow!10, 
    breakable, 
    boxrule=0pt, 
    arc=0pt, 
    left=5pt, right=5pt, top=5pt, bottom=5pt 
]

\begin{enumerate}[leftmargin=*, itemsep=0.5em]
    \item \textit{Prompt 1}: "Generate potential future climate policy scenarios for Sub-Saharan Africa that focus on accelerating renewable energy adoption. Consider policy incentives, grid integration challenges, and mechanisms for ensuring equitable access to clean energy."
    \item \textit{Prompt 2}: "Develop climate policy scenarios in which Sub-Saharan Africa prioritizes energy efficiency as a pillar of sustainable development. Include regulatory frameworks, financial incentives, and projected long-term benefits."
    \item \textit{Prompt 3}: "Propose future policy pathways that drive equitable access to renewable energy solutions in off-grid communities across Sub-Saharan Africa. Explore community-led initiatives, financing models, and regulatory support."
    \item \textit{Prompt 4}: "Develop climate policy scenarios where Sub-Saharan Africa prioritizes green hydrogen as a key energy export. Explore how governments might incentivize production, build supporting infrastructure, and navigate geopolitical opportunities and risks in global energy markets."
    \item \textit{Prompt 5}: "Imagine future climate policy scenarios where Sub-Saharan Africa invests heavily in research and development of new climate technologies. Highlight key focus areas, potential breakthroughs, and the role of international collaboration."
    \item \textit{Prompt 6}: "Generate scenarios in which artificial intelligence (AI) and digital technologies are integrated into climate policy implementation across Sub-Saharan Africa. Consider applications in emissions monitoring, predictive modeling, and adaptive policymaking."
    \item \textit{Prompt 7}: "Explore different climate policy scenarios in which Sub-Saharan Africa reduces fossil fuel dependency and transitions to a clean energy economy. Consider pathways that emphasize economic diversification, workforce retraining initiatives, and regulatory reforms."
    \item \textit{Prompt 8}: "Develop climate policy scenarios in which Sub-Saharan Africa adopts carbon pricing and emissions trading as primary tools for reducing greenhouse gas emissions. Consider potential economic benefits, implementation hurdles, and regional cooperation models."
    \item \textit{Prompt 9}: "Explore policy scenarios where Sub-Saharan Africa integrates circular economy principles into energy policy to promote sustainability. Discuss waste-to-energy innovations, material reuse, and industrial symbiosis."
    \item \textit{Prompt 10}: "Explore viable climate policy scenarios in which Sub-Saharan Africa adopts nuclear energy as part of its energy transition. Discuss key regional and international partnerships (e.g., IAEA, COP agreements), infrastructure development needs, and financial feasibility."
    \item \textit{Prompt 11}: "Develop climate policy scenarios for reducing transportation-related emissions in Sub-Saharan Africa. Discuss policy incentives for electric vehicle adoption, challenges in urban transport planning, and the role of biofuels and hydrogen as alternative energy sources."
    \item \textit{Prompt 12}: "Imagine future scenarios where decentralized energy efficiency initiatives significantly reduce energy poverty in Sub-Saharan Africa. Highlight local innovations, policy enablers, and community-driven approaches."
    \item \textit{Prompt 13}: "Develop potential future climate policy scenarios for Sub-Saharan Africa that emphasize sustainable land use and afforestation. Consider incentives for reforestation, carbon sequestration policies, and indigenous land management practices."
    \item \textit{Prompt 14}: "Generate possible future scenarios where Sub-Saharan Africa accelerates climate finance through international funding mechanisms. Consider public-private partnerships, sovereign green bonds, and risk mitigation strategies."
    \item \textit{Prompt 15}: "Propose detailed policy scenarios for Sub-Saharan Africa that strengthen institutional capacity for climate adaptation. Address governance reforms, inter-agency coordination, and financial resource allocation."
    \item \textit{Prompt 16}: "Explore future policy pathways where Sub-Saharan Africa mainstreams climate adaptation into national development plans. Discuss cross-sectoral collaboration, funding integration, and legislative frameworks."
    \item \textit{Prompt 17}: "Explore policy pathways where indigenous knowledge and local community-led solutions shape climate adaptation strategies in Sub-Saharan Africa. Consider land rights, traditional resource management practices, and equitable governance structures."
    \item \textit{Prompt 18}: "Generate policy scenarios where empowering local communities through climate education and advocacy leads to stronger grassroots climate action. Include capacity-building initiatives, knowledge-sharing platforms, and policy uptake metrics."
    \item \textit{Prompt 19}: "Imagine future climate policy scenarios where capacity building for climate governance strengthens institutional responses to climate challenges in Sub-Saharan Africa. Consider skills development, stakeholder collaboration, and policy implementation."
    \item \textit{Prompt 20}: "Please generate potential future policy scenarios for Sub-Saharan Africa that focus on building climate-resilient infrastructure—particularly around water management and disaster risk reduction. Include plausible timeframes, critical actors, and key funding mechanisms."
    \item \textit{Prompt 21}: "Propose detailed policy scenarios for Sub-Saharan Africa focusing on the health sector's adaptation to climate-induced challenges, such as heatwaves, vector-borne diseases, and flood-related health crises. Include metrics and expected outcomes."
    \item \textit{Prompt 22}: "Generate possible future scenarios in which Sub-Saharan Africa accelerates green finance for large-scale renewable energy projects. Consider international donors, private capital, and novel funding mechanisms, and explain major opportunities and risks."
    \item \textit{Prompt 23}: "What are potential future climate policy pathways in Sub-Saharan Africa if carbon markets and offset schemes become mainstream? Outline how governments, regional bodies, and local communities might participate or benefit."
    \item \textit{Prompt 24}: "Imagine several policy scenarios in which external technology transfer accelerates. How might local institutions be strengthened to adopt, maintain, and innovate on climate-related technologies (e.g., solar, wind, climate-smart agriculture)?"
    \item \textit{Prompt 25}: "Generate a set of climate policy scenarios for Sub-Saharan Africa that show how collaboration between local, national, and regional bodies might evolve. Include considerations of political will, stakeholder conflicts, and resource allocation."
    \item \textit{Prompt 26}: "Develop future policy pathways in which local communities and grassroots movements play a pivotal role in shaping national climate strategies. Consider equity issues, social inclusion, and mechanisms for ensuring marginalized groups have a voice."
    \item \textit{Prompt 27}: "Propose climate policy scenarios addressing transboundary resource management (e.g., shared water basins, pastoral lands) in Sub-Saharan Africa. Explore how climate stressors could exacerbate conflict or incentivize deeper regional cooperation."
    \item \textit{Prompt 28}: "Please propose climate policy scenarios that prioritize mitigation efforts with the largest public-health co-benefits, such as reducing indoor air pollution from traditional biomass cooking. Include measurable outcomes and ethical considerations."
    \item \textit{Prompt 29}: "Describe a range of worst-case ‘business as usual’ climate policy scenarios for Sub-Saharan Africa. Consider political inertia, minimal international support, and accelerating climate impacts, and explore the long-term social and economic consequences."
    \item \textit{Prompt 30}: "Generate scenarios in which Sub-Saharan Africa experiences more extreme climate impacts than currently predicted. How might governments, communities, and private actors innovate or pivot policy approaches in this high-risk future?"
    
\end{enumerate}

\textbf{Follow-up prompts}

\begin{enumerate}[leftmargin=*, itemsep=0.5em]

    \item \textit{Prompt 1}: What specific metrics can be used to measure the success of each scenario? Identify Key Performance Indicators (KPIs) that track progress and milestones that indicate meaningful advancements.
    \item \textit{Prompt 2}: Identify the key drivers that will influence the success or failure of each scenario. Consider factors such as economic conditions, policy frameworks, technology readiness, public perception, and environmental constraints. Additionally, analyze how these key drivers interact with each other, highlighting dependencies or feedback loops that could impact the scenario.
    \item \textit{Prompt 3}: Outline a clear implementation roadmap with well-defined milestones. Define short-term (0-2 years), medium-term (3-7 years), and long-term (8+ years) milestones. Highlight critical decision points, dependencies, and risks that could impact progress. Provide specific action steps that should be taken at each milestone to ensure scenario success.
    
\end{enumerate}
\end{tcolorbox}

\section{Response sample (rejected and valid response with validator notes)}
\label{app:scenarios}

\begin{tcolorbox}[
    colback=lightgray, 
    colframe=lightgray, 
    breakable, 
    boxrule=0pt, 
    arc=0pt, 
    left=5pt, right=5pt, top=5pt, bottom=5pt 
]

\subsection*{B.1    Rejected prompt and response}

\bigskip

\textbf{Prompt}: Please generate potential future climate policy scenarios for Sub-Saharan Africa related to renewable energy adoption.
\bigskip

\textbf{LLM response:}

\bigskip

\textbf{Scenario 1: "African Renaissance"}

\begin{itemize}[leftmargin=*, itemsep=0.5em]
    \item By 2030, Sub-Saharan Africa has become a global leader in renewable energy adoption, with a focus on solar and wind power.
    \item The African Union has set a target of generating at least 30\% of its electricity from renewable sources by 2030.
    \item Countries like South Africa, Nigeria, and Egypt have invested heavily in renewable energy infrastructure, including large-scale solar farms and wind parks.
    \item The region has also seen significant growth in decentralized renewable energy systems, such as rooftop solar and mini-grids, which have improved energy access for rural communities.
    \item International cooperation and investment have played a crucial role in supporting this transition, with organizations like the African Development Bank and the European Union providing significant funding and technical assistance.
\end{itemize}

\textbf{Scenario 2: "Climate-Resilient Economy"}
\begin{itemize}[leftmargin=*, itemsep=0.5em]
    \item By 2030, Sub-Saharan Africa has made significant progress in transitioning to a low-carbon economy, but faces ongoing challenges related to climate change.
    \item The region has implemented policies aimed at promoting renewable energy adoption, such as tax incentives and feed-in tariffs, but these have been unevenly applied across countries.
    \item Climate-resilient infrastructure development has become a priority, with investments focused on sea walls, flood protection systems, and green roofs to mitigate the impacts of climate change.
    \item However, the region still struggles with energy access and affordability, particularly in rural areas, where many communities rely on fossil fuels for heating and cooking.
    \item International cooperation is critical to supporting this transition, but also faces challenges related to funding and capacity building.
\end{itemize}

\textbf{Scenario 3: "Energy Poverty Trap"}

\begin{itemize}[leftmargin=*, itemsep=0.5em]
    \item By 2030, Sub-Saharan Africa has failed to make significant progress in transitioning to a low-carbon economy, with energy poverty remaining a major challenge.
    \item The region's reliance on fossil fuels has contributed to high greenhouse gas emissions and climate change impacts, including more frequent droughts and heatwaves.
    \item Energy access remains a major issue, particularly for rural communities, where many rely on expensive and polluting diesel generators or charcoal for heating and cooking.
    \item International cooperation has been limited, with developed countries prioritizing their own energy security interests over support for Sub-Saharan Africa's transition to renewable energy.
    \item The region faces significant challenges in building the capacity and institutions needed to support a low-carbon economy, including a lack of technical expertise and infrastructure.
\end{itemize}

\bigskip

\textbf{Validator notes:} The generated scenarios are unrealistic. This is largely because they are "negatives". For example, "Energy poverty trap" would not be a policy scenario but rather it is something that would happen as a result of lack of policy or policy execution. It seems the model assessed (llama3.2-3B) isn't able to distinguish the difference from the zero shot question. Perhaps it is not an issue of model strength but one of prompting approach.

\bigskip

\subsection*{B.2    Validated prompt and response}

\bigskip

\textbf{Prompt}: Generate potential future climate policy scenarios for Sub-Saharan Africa that focus on accelerating renewable energy adoption. Consider policy incentives, grid integration challenges, and mechanisms for ensuring equitable access to clean energy.

\bigskip

\textbf{LLM response:}

\bigskip

\textbf{Scenario 1: "African Renaissance"}

\begin{itemize}[leftmargin=*, itemsep=0.5em]
    \item \textbf{Policy Incentives:} Implement a carbon pricing mechanism (e.g., carbon tax or cap-and-trade system) to provide a financial incentive for renewable energy investments. Offer subsidies and grants for small-scale renewable energy projects, particularly in rural areas.
    \item \textbf{Grid Integration Challenges:} Invest in grid modernization and expansion to accommodate increasing renewable energy capacity. Develop smart grid technologies to manage variable renewable energy sources (e.g., solar and wind).
    \item \textbf{Equitable Access Mechanisms:}
    \begin{itemize}[leftmargin=*]
        \item Implement a national renewable energy target (e.g., 30\% of total electricity generation from renewables by 2030) with a focus on rural electrification.
        \item Establish a community-based renewable energy fund to support local communities in developing their own renewable energy projects.
        \item Develop a grid-scale energy storage system to ensure reliable power supply during periods of low renewable energy production.
    \end{itemize}
\end{itemize}

\textbf{Scenario 2: "African Energy Transition"}

\begin{itemize}[leftmargin=*, itemsep=0.5em]
    \item \textbf{Policy Incentives}: Offer tax credits and investment incentives for large-scale renewable energy projects, particularly in the power sector. Implement a feed-in tariff (FIT) scheme to provide a stable revenue stream for renewable energy generators.
    \item \textbf{Grid Integration Challenges:} Invest in advanced grid management systems to manage the variable output of renewable energy sources. Develop a national energy storage strategy to mitigate intermittency issues.
    \item \textbf{Equitable Access Mechanisms:}
    \begin{itemize}[leftmargin=*]
	\item Implement a national renewable energy policy framework that prioritizes rural electrification and energy access for vulnerable populations.
	\item Establish a Renewable Energy Development Bank to provide financing for renewable energy projects, particularly in underserved regions.
	\item Develop a peer-to-peer energy trading platform to enable households and businesses to buy and sell excess energy.
    \end{itemize}
\end{itemize}

\textbf{Scenario 3: "African Green Economy"}

\begin{itemize}[leftmargin=*, itemsep=0.5em]
    \item \textbf{Policy Incentives:} Implement a green economy framework that integrates climate change mitigation and adaptation with sustainable development goals. Offer subsidies and grants for small-scale renewable energy projects, particularly in the agriculture and water sectors.
    \item \textbf{Grid Integration Challenges:} Invest in decentralized energy systems (e.g., mini-grids, community-based energy systems) to provide reliable power supply to remote communities. Develop a national energy efficiency program to reduce energy demand.
    \item \textbf{Equitable Access Mechanisms:}
    \begin{itemize}[leftmargin=*]
	\item Implement a national renewable energy target with a focus on promoting energy access for marginalized communities.
	\item Establish a Community-Based Renewable Energy Fund to support local communities in developing their own renewable energy projects.
	\item Develop a climate-resilient agriculture program that integrates renewable energy and sustainable agricultural practices.
    \end{itemize}
\end{itemize}

\bigskip

\textbf{Validator notes:}  This is much better, including the responses to the re-formulated prompt. The generated scenarios align with the regional objectives on renewable energy targets and initiatives. The funding mechanism is plausible, and the focus on microgrids addresses infrastructure challenges in SSA.
\end{tcolorbox}

\section{Evaluation criteria for human climate expert ratings}
\label{app:criteria}

\begin{tcolorbox}[
    colback=lightgray, 
    colframe=lightgray, 
    breakable, 
    boxrule=0pt, 
    arc=0pt, 
    left=5pt, right=5pt, top=5pt, bottom=5pt 
]
This appendix outlines the key criteria used by the human climate evaluator to evaluate the generated responses. The criteria assess three dimensions: Faithfulness (A), Answer Relevancy (B), and Context Utilization (C), rated on a scale from 1 (Very Poor) to 5 (Excellent).

\begin{itemize}[leftmargin=*, itemsep=0.5em]
    \item \textbf{1 — Very Poor}
    \begin{itemize}[leftmargin=*]
        \item \textbf{A: Faithfulness} The answer significantly contradicts or ignores the referenced context, introduces major factual errors or “hallucinations,” or cannot be justified by the source material (e.g., COP reports or SSA policy data).
        \item \textbf{B: Answer Relevancy} The answer fails to address the question’s main points, strays off-topic, or provides irrelevant information.
        \item \textbf{C: Context Utilization} The answer barely uses the retrieved context or uses none at all and/or includes large amounts of irrelevant information.
    \end{itemize}

    \item \textbf{2 — Poor}
    \begin{itemize}[leftmargin=*]
        \item \textbf{A: Faithfulness} The answer partially aligns with the references but contains noticeable inaccuracies or unsupported claims.
        \item \textbf{B: Answer Relevancy} The answer touches on the question but omits key aspects or provides minimal relevant detail.
        \item \textbf{C: Context Utilization} The answer shows only a limited or superficial link to the retrieved context, with more irrelevant than relevant content.
    \end{itemize}

    \item \textbf{3 — Moderate / Fair}
    \begin{itemize}[leftmargin=*]
        \item \textbf{A: Faithfulness} The answer does not outright contradict the context, though it may include some unverified extrapolations or minor inconsistencies.
        \item \textbf{B: Answer Relevancy} The answer addresses the main question to a reasonable extent but may lack depth or clarity in some areas.
        \item \textbf{C: Context Utilization} The answer uses the retrieved context in a broad or generic way, with a roughly equal mix of relevant and possibly extraneous details.
    \end{itemize}

    \item \textbf{4 — Good}
    \begin{itemize}[leftmargin=*]
        \item \textbf{A: Faithfulness} The answer is solidly grounded in the references, generally accurate, and avoids significant factual errors or unjustified claims.
        \item \textbf{B: Answer Relevancy} The answer thoroughly addresses the question, providing clear and direct responses to the key points asked for.
        \item \textbf{C: Context Utilization} The answer makes effective use of the retrieved context, selectively incorporating relevant details without much irrelevant information.
    \end{itemize}

    \item \textbf{5 — Excellent}
    \begin{itemize}[leftmargin=*]
        \item \textbf{A: Faithfulness} The answer is fully consistent with the source material, highly accurate, and all major claims are justified by the references.
        \item \textbf{B: Answer Relevancy} The answer precisely targets the question, comprehensively covering all important aspects in a well-structured manner.
        \item \textbf{C: Context Utilization} The answer draws deeply and accurately on the retrieved context, weaving in the most pertinent points while avoiding extraneous detail.
    \end{itemize}
\end{itemize}
\end{tcolorbox}

\end{document}